\pdfoutput=1
\documentclass[journal,twoside,web]{ieeecolor}
\usepackage{tmi}
\usepackage{cite}
\usepackage{amsmath,amssymb,amsfonts}
\usepackage{graphicx}
\usepackage{textcomp}
\def\BibTeX{{\rm B\kern-.05em{\sc i\kern-.025em b}\kern-.08em
    T\kern-.1667em\lower.7ex\hbox{E}\kern-.125emX}}
    
\usepackage{bbm}
\usepackage{ctable}
\usepackage{multirow}
\usepackage{subcaption}
\usepackage[normalem]{ulem}
    
\graphicspath{{figures/pdf/}{figures/png/}}
\DeclareGraphicsExtensions{.pdf,.png}

\newcommand{\mcc}[1]{\multicolumn{1}{c}{#1}} 
\newcommand{\etal}{\mbox{\emph{et al.}}}
\newcommand{\sota}{\mbox{state-of-the-art}}

\begin{document}
\title{Cardiac Segmentation with Strong Anatomical Guarantees}
\author{Nathan~Painchaud,
        Youssef~Skandarani,
        Thierry~Judge,
        Olivier~Bernard,
        Alain~Lalande,
        and~Pierre-Marc~Jodoin \vspace{-0.5cm}
\thanks{Manuscript received April 12, 2020; revised June 10, 2020.}
\thanks{N. Painchaud, T. Judge and P.-M. Jodoin are with the Department
of Computer Science, University of Sherbrooke, Sherbrooke,
QC, Canada (e-mail: nathan.painchaud@usherbrooke.ca).}
\thanks{A. Lalande is with University of Bourgogne Franche-Comte, Dijon, France.}
\thanks{Y. Skandarani is with University of Bourgogne Franche-Comte and CASIS inc. Dijon, France.}
\thanks{O. Bernard is with University of Lyon, CREATIS, CNRS UMR5220, Inserm U1206, INSA-Lyon, University of Lyon 1, Villeurbanne, France.}
\thanks{Copyright \textcopyright 2020 IEEE. Personal use of this material is permitted. However, permission to use this material for any other purposes must be obtained from the IEEE by sending a request to pubs-permissions@ieee.org.}}

\maketitle

\begin{abstract}
 Convolutional neural networks (CNN) have had unprecedented success in medical imaging and, in particular, in medical image segmentation.  However, despite the fact that segmentation results are closer than ever to the inter-expert variability, CNNs are not immune to producing anatomically inaccurate segmentations, even when built upon a shape prior.  In this paper, we present a framework for producing cardiac image segmentation maps that are guaranteed to respect pre-defined anatomical criteria, while remaining within the inter-expert variability. The idea behind our method is to use a well-trained CNN, have it process cardiac images, identify the anatomically implausible results and warp these results toward the closest anatomically valid cardiac shape.  This warping procedure is carried out with a constrained variational autoencoder (cVAE) trained to learn a representation of valid cardiac shapes through a smooth, yet constrained, latent space.  With this cVAE, we can project any implausible shape into the cardiac latent space and steer it toward the closest correct shape.  We tested our framework on short-axis MRI as well as apical two and four-chamber view ultrasound images, two modalities for which cardiac shapes are drastically different.
 With our method, CNNs can now produce results that are both within the inter-expert variability and always anatomically plausible without having to rely on a shape prior.
\end{abstract}


\section{Introduction}
\IEEEPARstart{M}{agnetic} Resonance Imaging (MRI) and ultrasound imagery (US) are the most widely-used cardiac image acquisition devices in clinical routine.  While MRI can produce high-contrast, high-resolution and high-SNR images in any orientation, the cardiac function is typically evaluated from a series of kinetic images (cine-MRI) acquired in short-axis orientation of the left ventricle~\cite{Salerno17}.  In clinical practice, cardiac parameters are usually estimated from the knowledge of the endocardial and epicardial borders of the left ventricle (defining the left cavity (LV) and the myocardium (MYO)) and the endocardial border of the right ventricle (RV) at the end-diastolic (ED) and end-systolic (ES) phases. MRI is the reference exam for the evaluation of the cardiac function and of the cardiac viability after myocardial infarction. Unfortunately, the MRI device is bulky, expensive and cannot be operated by one person even with the latest innovations.

On the other hand, echocardiography is an highly flexible and low-cost exam to evaluate the cardiac function. Ultrasound devices are small and less expensive that one can carry around the hospital.  As such, US provides physicians real-time images in an easy way and is often described as the modern stethoscope.  Unfortunately, ultrasound images suffer from a poor SNR, noise artifacts, local signal drop, limited field of view, and a limited acquisition angle.  The most widely-used acquisition protocol to evaluate the cardiac function involve a 2D+time long-axis orientation resulting into two and four-chamber view images.  As for MRI, the endocardial and epicardial borders are outlined at the ED and ES time instant.  The volume and ejection fraction of the LV is then computed with the biplane Simpson's formula~\cite{Folland79SimpsonEF}.

US and MRI are complementary by nature.  US devices can quickly evaluate the heart function, find the source of certain symptoms and detect or exclude pathologies. MRI is an imaging modality to further assess a disease and for longitudinal analysis.  Both MRI and US are non-invasive and are non-irradiating imaging techniques.

CNNs have had great success at segmenting these  modalities~\cite{Duan19,Bernard2018DeepLT,Leclerc19,Oktay17,Zotti18}.  Some neural nets even provide results with overall Dice index and/or Hausdorff distance within the inter and intra-observer variations~\cite{Bernard2018DeepLT,Leclerc19}. Unfortunately, these methods still generate spurious anatomically impossible shapes with holes inside the structures, abnormal concavities, and duplicated regions to name a few. Therefore, despite their excellent results on average, these methods are still unfit for a day-to-day clinical use.

To reduce such errors, some authors integrate shape priors to their model~\cite{Duan19,Oktay17,Zotti18} while others simply post-process the generated shapes with morphological operators or some connected component analysis to remove small isolated regions. Unfortunately, none of these approaches can guarantee 100\% of the time the anatomical plausibility of their results.

In this paper, we present the first deep learning formalism which guarantees the anatomical plausibility of cardiac shapes, w.r.t. well-defined criteria, under any circumstances. Our method can be plugged at the output of any segmentation method to reduce to zero its number of anatomically invalid shapes, while preserving its overall accuracy.  As will be shown in the results section, the same framework is effective for a variety of segmentation methods both applied on echocardiographic and MR images.

\section{Previous Work}
Although there is more non-deep-learning cardiac segmentation methods than deep learning ones (neural networks are relatively new in the field) we shall focus on the latter due to the very nature of our contribution.

\begin{figure*}[tp]
\centering
\includegraphics[width=0.96\textwidth]{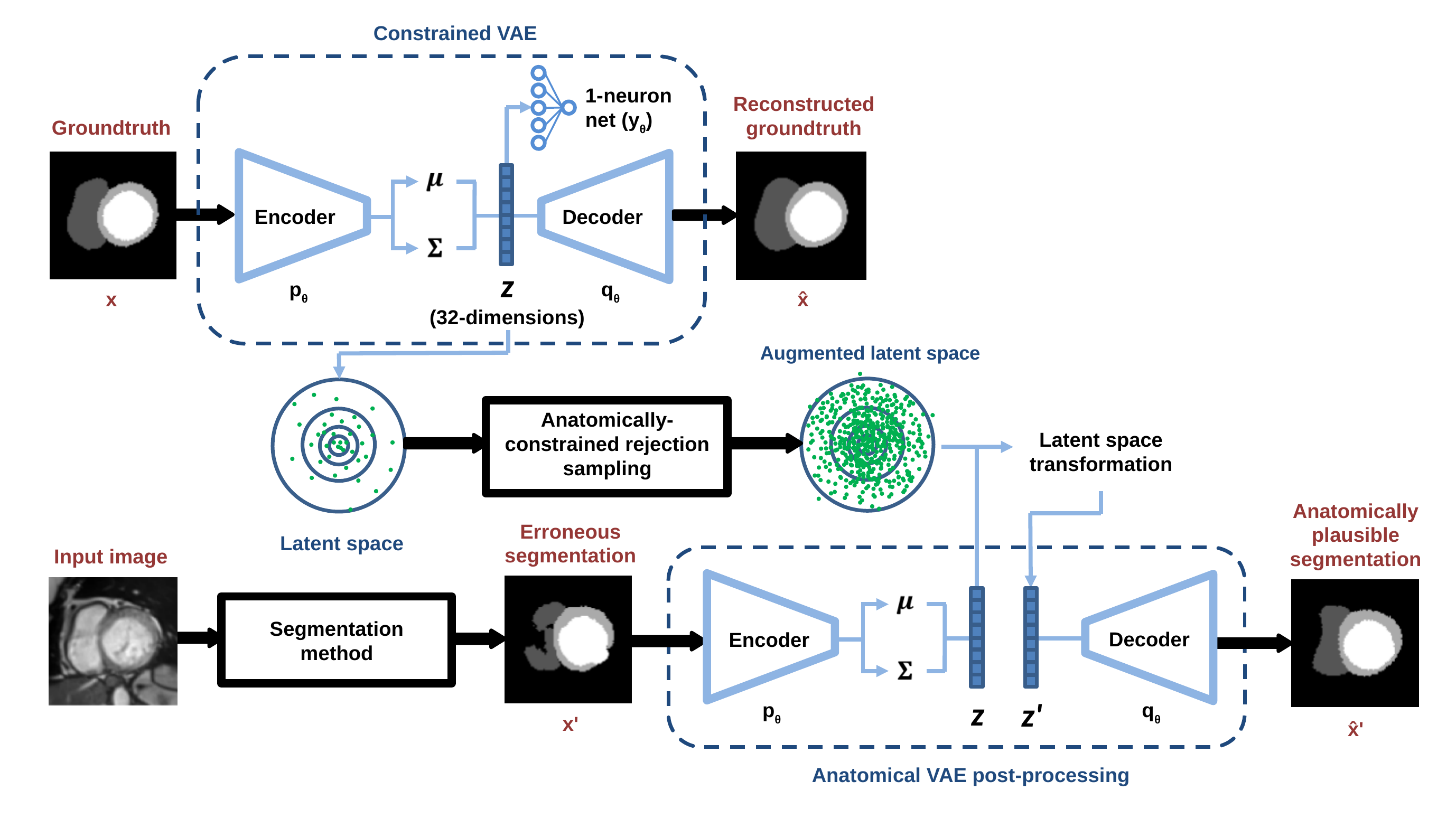}\vspace{-0.2cm}
\caption{Schematic representation of our method.  Although this figure illustrates short axis image segmentation, we use the same framework for two and four-chamber view ultrasound images.  The segmentation method is a placeholder for any cardiac segmentation method. The parameters used by the anatomical VAE ($p_\theta,q_\theta$) are the same ones trained on the constrained VAE. \vspace{-0.3cm} }
\label{fig:ourmethod}
\end{figure*}

\vspace{-0.3cm}
\subsection{MRI segmentation}
\subsubsection*{CNNs} The U-Net~\cite{Ronneberger2015UNet} has become the {\em de facto} generic encoder-decoder CNN for biomedical image segmentation and is often used in cardiology. Isensee \mbox{\emph{et al.}}~\cite{isensee2017automatic}, winner of the 2017 MICCAI Automated Cardiac Diagnosis Challenge (ACDC)~\cite{Bernard2018DeepLT}, used an ensemble of 2D and 3D U-Net, with the addition of an upscaling and aggregation of the last two convolutional blocks of the decoder for the final segmentation.  Also, as mentioned by Bernard \mbox{\emph{et al.}}~\cite{Bernard2018DeepLT}, several other challengers used a modified version of the UNet.
Vigneault \mbox{\emph{et al.}} proposed a more domain specific approach, Omega-Net~\cite{vigneault2018omega}, which has, at its heart, a localization and transformation network that transforms the input MRI into a canonical orientation which is subsequently segmented by a cascade of U-Nets. 

\subsubsection*{CNNs with shape prior}
Although most deep segmentation methods produce accurate segmentation results, they still suffer from anatomical inconsistencies.  As a solution, several authors incorporate a shape prior to their model.
Oktay \mbox{\emph{et al.}} uses an approach named {\em anatomically constrained neural network} (ACNN)~\cite{Oktay17}. Their neural network is similar to a 3D U-Net whose segmentation output is constrained to be close to a non-linear compact representation of the underlying anatomy, derived from an autoencoder network. More recently, Zotti \mbox{\emph{et al.}} proposed a method based on the grid-net architecture that embeds a cardiac shape prior to segment MR images~\cite{Zotti18}. Their shape prior encodes the probability of a 3D location point being a member of a certain class and is automatically registered with the last feature maps of their network. Finally, Duan \mbox{\emph{et al.}} implemented a shape-constrained bi-ventricular segmentation strategy \cite{Duan19}. Their pipeline starts with a multi-task deep learning approach that aims to locate specific landmarks. These landmarks are then used to initialize atlas propagation during a refinement stage of segmentation. Although the use of an atlas improves the quality of the results, their final segmented shapes strongly depend on the accuracy of the located landmarks. From these studies, it appears that only soft constraints are currently imposed in the literature to steer the segmentation outputs toward a reference shape.  As will be shown in this paper, shape-prior methods are not immune to producing anatomically incorrect results.

\vspace{-0.3cm}
\subsection{Echocardiographic segmentation}

\subsubsection*{CNNs}
In 2012, Carneiro \etal~exploited deep belief networks and the decoupling of rigid and nonrigid classifiers to improve robustness in terms of image conditions and shape variability~\cite{Carneiro2012}. Later, Chen \etal~used transfer learning from cross domain to enhance feature representation~\cite{Chen2016}. In parallel, Smistad \etal~showed that the \mbox{U-Net}~\cite{Ronneberger2015UNet} could be trained with the output of a \sota~deformable model to segment the LV in 2D ultrasound images \cite{Smistad2017}. Additionally, Leclerc \etal~showed that a simple \mbox{U-Net} learned from a large annotated dataset can produce accurate results that are much better than the \sota, on average below the inter-observer variability and close but still above the intra-observer variability with 18\% of outliers \cite{Leclerc19}. Recently, the same authors proposed to efficiently integrate the U-Net into a multi-task network (the so-called "LUNet") designed to optimize in parallel a localization and a segmentation procedure~\cite{Leclerc2020Lunet}. Their results showed that localization allows the introduction of contextualization properties which improve the  overall accuracy of cardiac segmentation while reducing the number of outliers to 11\%.

\subsubsection*{CNNs with shape prior}
The ACNN model proposed by Oktay \etal~\cite{Oktay17} was also applied to the segmentation of the endocardial border in 3D echocardiography. Results showed that the use of an autoencoder network to impose soft shape constraints allowed to obtain highly competitive scores with respect to the \sota~while learning from a limited number of cases (30 annotated volumes). Very recently, Dong \etal~developed a deep atlas network to significantly improve 3D LV segmentation based on limited annotation data~\cite{Dong2020}. The key aspects of this architecture are a light-weight network to perform registration and a multi-level information consistency constraint to enhance the overall model's performance. This method currently has the best scores for 3D LV segmentation in 3D echocardiography.  Jafari \etal~also proposed to alter the echocardiography fed to segmentation models using a framework that introduces soft shape priors to Cycle-Gan~\cite{Jafari19}. By enhancing the quality of the input images through image translation, the authors manage to improve the worst-case performance of standard segmentation networks.

\vspace{-0.3cm}
\section{Proposed Framework}
\vspace{-0.2cm}
A schematic representation of our method is given in Fig.~\ref{fig:ourmethod}. The system is used for both short-axis MR images and long-axis echocardiographic images, two fairly different looking cardiac shapes.  Overall, the system is made of three blocks, namely: 1) a constrained VAE that learns the latent representation of valid cardiac shapes, 2) an anatomically-constrained rejection sampling procedure to augment the number of latent vectors and 3) a post-processing VAE that warps anatomically invalid shapes toward the closest valid ones.  Since the system implements a post-processing for segmentations, the "Segmentation method" block in Fig.~\ref{fig:ourmethod} is a placeholder for any possible cardiac segmentation method.  The anatomical guarantees come from an operation called "Latent space transformation" in Fig.~\ref{fig:ourmethod}, that substitutes the latent vector of an incorrect shape by a close but valid one.

The correctness of a cardiac shape is determined by a set of complementary anatomical criteria.  These criteria allow to identify anatomically implausible configurations regardless of the input image.  As such, the aim of our system is to output cardiac shapes that always respect these anatomical criteria.

\vspace{-0.2cm}
\subsection{Anatomical Criteria}\label{sec:criteria}
\begin{figure}[tp]
  \begin{subfigure}[b]{0.495\columnwidth}
    \includegraphics[width=\textwidth]{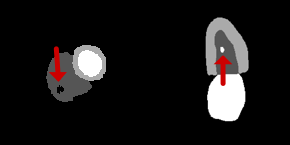}
    \caption{Intra-structure holes}
  \end{subfigure}
  \hfill
  \begin{subfigure}[b]{0.495\columnwidth}
    \includegraphics[width=\textwidth]{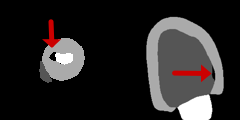}
    \caption{Inter-structure holes}
  \end{subfigure}
  \vfill
  \begin{subfigure}[b]{\columnwidth}
    \includegraphics[width=\textwidth]{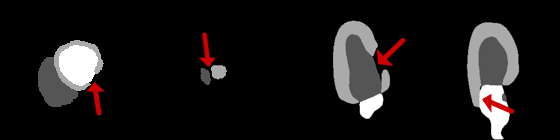}
    \caption{Connectivity and LV cavity touching background}
  \end{subfigure}
  \vfill
  \begin{subfigure}[b]{0.495\columnwidth}
    \includegraphics[width=\textwidth]{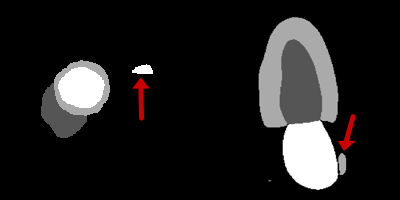}
    \caption{Fragmented structures}
  \end{subfigure}
  \hfill
  \begin{subfigure}[b]{0.495\columnwidth}
    \includegraphics[width=\textwidth]{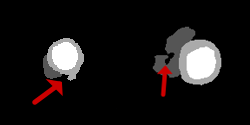}
    \caption{Concavity}
  \end{subfigure}
  \caption{\small Examples of anatomically erroneous cardiac shapes for both short and long-axis views.}
  \label{fig:anatomical_errors_examples}
\end{figure}

\vspace{-0.0cm}
Because of the orientation used to acquire cine MR and apical ultrasound images, our system uses two sets of anatomical criteria, namely the short-axis and the long-axis criteria (c.f.~Fig.~\ref{fig:anatomical_errors_examples} and ~\ref{fig:generated_samples} for illustrations).  When relevant, thresholds were defined based on the datasets' training set (ACDC for short-axis, CAMUS for long-axis) so that no clinically relevant segmentations were marked as invalid. Both datasets cover healthy and pathological cases, so  the thresholds take into account a representative distribution of cardiac configurations, and not only a subset of healthy configurations. Since these criteria are not included in the loss, they do not need to be differentiable.  They are evaluated systematically on every sample from the latent space, so they do however need to be computable algorithmically using traditional image processing, for efficiency concerns.

\subsubsection*{Short-Axis Criteria}\label{sec:criteria:sa}
Our system uses 16 anatomical short-axis criteria that each highlight an invalid cardiac configuration. These criteria are the following:

\begin{enumerate}
    \item ({\em 3 criteria}) hole(s) in the LV, the RV or the MYO
    \item ({\em 2 criteria}) hole(s) between the LV and the MYO and between the RV and the MYO
    \item ({\em 3 criteria}) the presence of more than one LV, RV or MYO
    \item ({\em 1 criterion}) the RV is disconnected from the MYO
    \item ({\em 2 criteria}) the LV touches the RV or the background
    \item ({\em 3 criteria}) the LV, RV and MYO has one (or more) acute concavity
    \item ({\em 2 criteria}) both for the LV and the MYO, the ratio of their area to that of a circle having the same perimeter (aka circularity metric) exceeds a certain threshold.
\end{enumerate}

\subsubsection*{Long-Axis Criteria}\label{sec:criteria:la}
We use 12 anatomical long-axis criteria to highlight invalid configurations.  These criteria are:

\begin{enumerate}
    \item ({\em 3 criteria}) hole(s) in the LV, MYO and left atrium (LA)
    \item ({\em 2 criteria}) hole(s) between the LV and the MYO or between the LV and the LA
    \item ({\em 3 criteria}) the presence of more than one LV, MYO or LA
    \item ({\em 2 criteria}) the size of the area by which the LV touches the background or the MYO touches the LA exceeds a certain threshold.
    \item ({\em 1 criterion}) the ratio between the minimal and maximal thickness of the MYO is below a given threshold
    \item ({\em 1 criterion}) the ratio between the width of the LV and the average thickness of the MYO exceeds a certain threshold. Both width and thickness are computed as the total width of the structure at the middle-point of the embeded bounding box. The goal is to identify situations for which the MYO is too thin with respect to the size of the LV.
\end{enumerate}

\subsection{Constrained Variational Autoencoder (cVAE)}
VAEs~\cite{kingma2013auto} are unsupervised neural networks trained to learn the latent representation of a set of data.  These neural nets are made of an encoder, which projects an input signal $\mathbf{x}$ to the latent space, and a decoder, which converts a latent vector $\vec z$ back into the input space.  More specifically, the VAE encoder outputs the parameters ($\vec \mu$ and $\Sigma$) of a Gaussian distribution $p_{\theta_e}(\vec z | \mathbf{x})$ where $\vec z \in \mathbb{R}^{k}$ is a latent vector ($k=32$ in our case, and $\theta_e$ are the parameters of the encoder network). The decoder takes in a latent variable $\vec z$ sampled from $p_{\theta_e}(\vec z | \mathbf{x})$ and outputs $\hat{\mathbf{x}}$, the reconstruction of the input vector $\mathbf{x}$.  As such, the decoder gets to learn the conditional distribution $q_{\theta_d}(\mathbf{x}|\vec z)$ with $\theta_d$ as the decoder parameters.

In this work, $\mathbf{x}$ and $\hat{\mathbf{x}}$ are 2D cardiac shapes, both $\in \mathbb{R}^{n\times n}$.  Since our overarching objective is to learn the latent representation of $valid$ cardiac shapes, we train our VAE with input values $\mathbf{x}$ that are groundtruth cardiac shapes outlined by a medical expert, and thus without any anatomical aberrations.  As such, after the VAE has been trained, Gaussian centroids $\vec \mu_i$ encoded from groundtruth cardiac shapes  $\mathbf{x}$ will also lead to an anatomically valid reconstructed shape $\hat{\mathbf{x}}$.  In fact, any point $\vec z$ sampled on the manifold of valid cardiac vectors can be decoded to an anatomically valid cardiac shape $\hat{\mathbf{x}}$.  As such, we call these vectors {\em valid} latent vectors.

However, as will be shown later, our method needs to linearly interpolate latent vectors.  It follows that a latent vector $\vec z$ interpolated between two anatomically valid vectors $\vec z_i,\vec z_j$ should also be valid (at least most of the time).  Furthermore, our method needs that a small translation $\vec \delta_z$ performed on a valid latent vector $\vec z$ leads to a smooth and anatomically coherent transformation on the resulting decoded image.  

These constraints can be fulfilled with a linear manifold that we approach with a constrained VAE (cVAE)~\cite{Higgins2017betaVAELB}.  The constraint comes in the form of a single-neuron regression network~\cite{Bishop07} $y_{\theta_c}(\vec z)$ trained simultaneously with the encoder and the decoder (c.f. Fig.~\ref{fig:ourmethod}).  The goal of the linear regression network is to reproduce a domain-specific target $t$ associated to the input image $\mathbf{x}$. Since a single-neuron network with no activation can only learn a linear function, the gradient from the regression loss forces the encoder to learn a more linear (and thus less convoluted) manifold of valid shapes in the latent space.  

The resulting loss function of our cVAE is:
\begin{eqnarray}
I\!\!E_{q_{\theta_e}(\vec z|\mathbf{x})}[-\log q_{\theta_d}(\mathbf{x}|\vec z)]+KL(p_{\theta_e}(\vec z|\mathbf{x}) \| p(\vec z)) + \\
\|y_{\theta_c}(\vec z)-t\|^2 \nonumber
\end{eqnarray}
where the first two terms make up the usual ELBO (Evidence Lower BOund) VAE loss function~\cite{kingma2013auto}, with $p(\vec z)$ as the unit-variance zero-mean Gaussian prior.  The last term is the L2 regression loss of the one-neuron net.

\subsubsection*{MRI short-axis linear constraint}
Since cine-MR short-axis images  $\mathbf{x}$ are 2D+time arrays stacked into two 3D volumes, in our study, only the ES and ED phases are considered and then the target predicted by the one-neuron regression network $y_{\theta_c}(\vec z)$ is the slice index of $\mathbf{x}$ normalized between 0 (base) and 1 (apex).

\subsubsection*{Ultrasound long-axis linear constraint}
The ultrasound signal is a 2D+time sequence of images.  In this case, the regression network is designed to predict the time instant of the input image $\mathbf{x}$.  Here as well, the target value is normalized between $0$ and $1$, where $0$ stands for the end-diastolic time instant and $1$, the end-systolic time instant.

\subsection{Anatomically-Constrained Data Augmentation}\label{sec:augmentation}
As mentioned before, once the cVAE is trained, the 2D groundtruth cardiac shapes $\mathbf{x}$ can be projected in the 32D latent space, where they form a manifold of {\em valid} latent vectors.  These latent vectors are "anatomically correct", since the deterministic cVAE decoder can convert them back to anatomically valid cardiac shapes.  

The idea behind our method is to warp invalid cardiac shapes toward a close but valid configuration.  This is done by projecting any invalid cardiac shape $\mathbf{x}$ to the latent space, project its associated {\em invalid} latent vector to the closest point on the manifold of {\em valid} latent vectors, and then decode the resulting vector.  Unfortunately, with 32 dimensions, the latent space has a whopping number of $2^{32}$ quadrants, which is orders of magnitude larger than any annotated cardiac dataset.  As such, with too few {\em valid} latent vectors, the manifold is too sparse to be effective.

One solution to that problem is to increase the number of {\em valid} latent vectors through data augmentation.  Since the manifold in the latent space is roughly linear, one can easily sample it with a rejection sampling (RS) method~\cite{Koller09}.  The goal is to generate a new set of latent vectors $Z'$ such that the distribution $P(\vec {z'})$ of these newly generated samples is close to $P(\vec z)$, the distribution from which the original {\em valid} latent vectors are identically independent and identically distributed (iid) from.  Since sampling $P(\vec z)$ directly is difficult, RS samples a second, and yet easier, probability density function $Q(\vec z)$.   A common choice for $Q(\vec z)$ is a Gaussian of mean and variance equal to the distribution of the original {\em valid} latent vector derived from the groundtruth segmentation.  A key idea with RS is that $P(\vec z) > MQ(\vec z)$ where $M > 1$.  Given $P(\vec z)$ and $Q(\vec z)$, the sampling procedure first generates a random sample $\vec z_j$, iid of $Q(\vec z)$, as well as a uniform random value $u\in [0,1]$.  If $u <  \frac{P(\vec z_j)}{M Q(\vec z_j)}$ then $\vec z_j$ is kept, otherwise it is rejected.  Since in our case $P(\vec z)$ is unknown {\em a priori}, we estimate it with a Parzen window distribution~\cite{Bishop07}.
 
\begin{figure}[tp]
\centering
  \begin{subfigure}[b]{\columnwidth}
    \includegraphics[width=\textwidth]{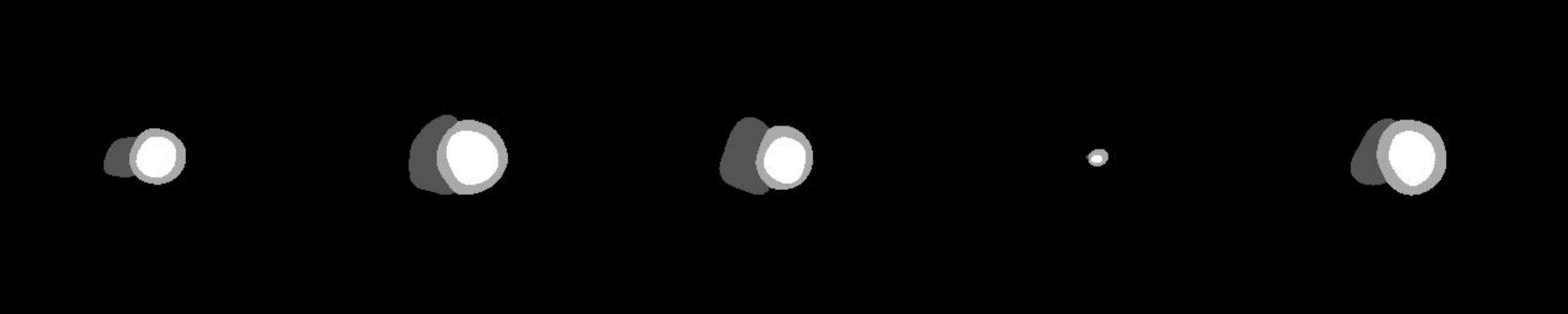}
    \caption{MRI (ACDC)}
  \end{subfigure}
  \vfill
  \begin{subfigure}[b]{\columnwidth}
    \includegraphics[width=\textwidth]{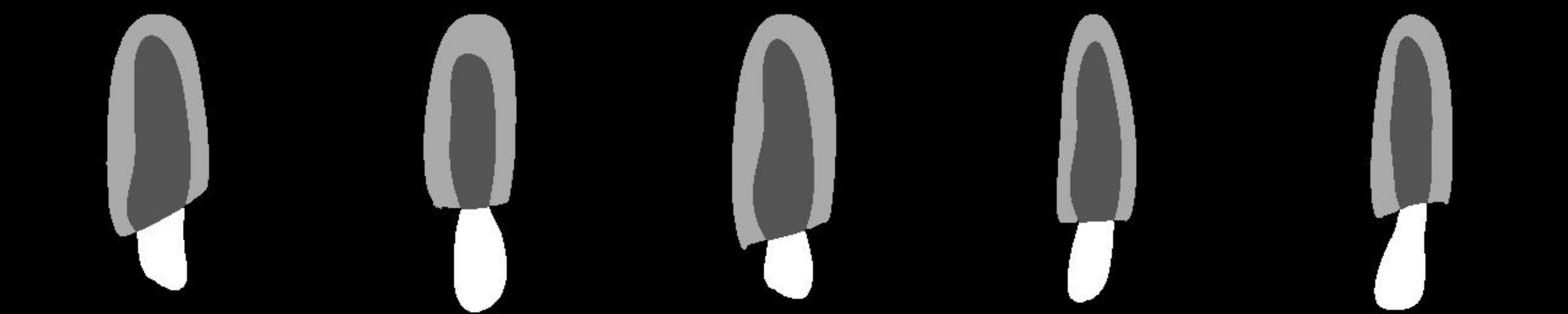}
    \caption{US (CAMUS)}
  \end{subfigure}
  \caption{\small Samples generated through anatomically-constrained data augmentation for MRI (short-axis view) and US (long-axis view).}
  \label{fig:generated_samples}
\end{figure}

The primary objective with RS is to increase the number of latent vectors.  However, since these newly generated points need to lie on the manifold of {\em valid} vectors, we want those new vectors to correspond to anatomically {\em valid} cardiac shapes.  As such, we redefine the RS criterion as follows:
\begin{eqnarray}
    u < \mathbbm{1}\left ( \mbox{dec}(\vec z_i) \right) \frac{P(\vec z_j)}{M Q(\vec z_j)}
\end{eqnarray}
where $\mbox{dec}(\vec z_j)$ is the VAE decoder that converts the latent vector $\vec z_j$ into a segmentation map and $\mathbbm{1}$ is an indicator function which returns 1 when the input segmentation map respects the defined anatomical criteria and zero otherwise.  In Fig.~\ref{fig:ourmethod}, this operation is called {\em anatomically-constrained rejection sampling augmentation}.  This sampling procedure is repeated up until the desired number of samples is reached.  At the end, a total number of 4 million latent vectors have been generated, both for the MRI and the ultrasound datasets.  Each of these vectors have a corresponding valid cardiac shape that respects the aforementioned anatomical criteria (c.f. Section~\ref{sec:criteria}).  Samples of cardiac shapes generated with anatomically-constrained rejection sampling augmentation are provided in Fig.~\ref{fig:generated_samples}.

\subsection{Cardiac shape warping}\label{sec:transformation}
Our system can be seen as a post-processing operator that one can plug after any segmentation method that sometimes generates anatomically erroneous segmentation maps.  This is illustrated at the bottom right of Fig.~\ref{fig:ourmethod}, where a VAE is used to convert erroneous segmentation maps into anatomically valid segmentations.  This post-processing VAE is in fact the trained cVAE.  Thus, any anatomically invalid segmentation map $\mathbf{x}$ fed to the VAE {\em encoder} gets projected into the latent space where 4 million {\em valid} vectors lie.  Furthermore, since the VAE {\em decoder} is deterministic, any anatomically valid latent vector $\vec z$ is guaranteed to be converted into an anatomically plausible cardiac shape.

As mentioned before, our aim is to warp an anatomically incorrect cardiac shape toward a close but correct configuration.  We do so by translating the latent vector $\vec z$ of an erroneous cardiac shape to a near but anatomically valid latent vector $\hat z$.  This operation can be summarized as:
\begin{eqnarray}
\label{eq:optim_nn}
\hat z_{\mbox{\tiny opt}}= \arg \min_{\hat z} ||\vec z - \hat z||^2, \,\,\,\, s.t.\,\, \mathbbm{1}\left ( \mbox{dec}(\hat z) \right)=1. 
\end{eqnarray}
The result of this optimization is a valid latent vector $\hat z$ that is the closest to $\vec z$.  However, since  $\mathbbm{1}\left ( \mbox{dec}(\hat z) \right)=1$ involves non-differentiable anatomical criteria, the optimization formulation of Eq.~(\ref{eq:optim_nn}) cannot be solved with a usual Lagrangian solution.  
An alternative solution is to redefine the problem of finding $\hat z$ as the problem of finding the smallest vector $\vec \delta_{z'}$ such that $\hat z=\vec z + \alpha \vec \delta_{z'}$ with $\alpha\in [0,1]$.
In our case, we recover $\vec \delta_{z'}$ based on the nearest neighbor of $\vec z$ in the augmented latent space, {\em i.e.} $\vec \delta_{z'} =(\vec z'_{N1}-  \vec z)$ where $\vec z'_{N1}$ corresponds to the nearest latent vector. This leads to an easier 1D optimization problem:
\begin{eqnarray}
\label{eq:optim_dicho}
\alpha_{\mbox{\tiny opt}} = \arg \min_{\alpha} |\alpha|, \,\,\,\, s.t. \,\,\mathbbm{1}\left ( \vec z + \alpha \vec \delta_{z'} \right)=1
\end{eqnarray}
that we solve with a dichotomic search.  Starting with $\alpha=0.5$, at each iteration, the anatomical criterion $\mathbbm{1}\left ( \mbox{dec}(\vec z + \alpha \vec \delta_{z'}) \right)$ dictates which half of the search space should be explored further: lower values of $\alpha$ if $\mathbbm{1}\left ( \mbox{dec}(\hat z) \right)=1$, and higher values of $\alpha$ if $\mathbbm{1}\left ( \mbox{dec}(\hat z) \right)=0$.

Since the dichotomic search reduces the search space exponentially fast, the optimization algorithm is stopped after five iterations.  At the end, the selected  $\alpha$ is the smallest that validates the anatomical criterion.

\begin{figure}[tp]
\centering
\includegraphics[width=0.95\linewidth]{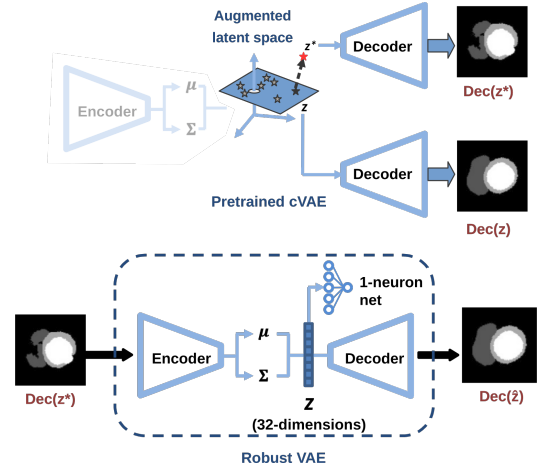}
\caption{\small [Top] Method used to generate invalid cardiac shape from valid ones. [Bottom] The valid/invalid pair of images are then used to train the Robust VAE.}
\label{fig:robustVAE}
\end{figure}

\subsection{Robust VAE}\label{sec:robustVAE}
Current limitations of the proposed method are the need for millions of latent vectors to be stored in memory and the nearest neighbor search to perform each time a segmentation result is anatomically flawed.  In this section, we present an alternative method that does not require the storage of latent vectors, nor the search for nearest neighbors.  This method allows for faster processing and reduced memory usage, but without the previous method's anatomical guarantees.  The use of either method depends on the application at hand.

Instead of using the post-processing VAE with cardiac shape warping as in Fig.~\ref{fig:ourmethod}, we implemented a {\em robust} VAE (rVAE). 
The goal of this new VAE is to directly convert erroneous segmentation maps $\mathbf{x}$ into anatomically plausible configurations $\hat{\mathbf{x}}$.  To do so, we added a step to the VAE training procedure. Starting with a pretrained cVAE, we fixed the weights of the decoder and of the single-neuron regression network, and fine-tuned the cVAE like a denoising autoencoder, i.e. by feeding it with anatomically implausible maps and training it to reproduce valid segmentations (c.f. bottom of Fig.~\ref{fig:robustVAE}). Since the decoder is fixed, this forces the new encoder of the cVAE ($\theta_e'$) to learn to project erroneous segmentation maps close to their corresponding valid latent vectors.

In practice, we generated a synthetic training set of 10,000 pairs of anatomically valid and invalid cardiac shapes, using the generative capabilities of the cVAE.  As shown at the top of Fig.~\ref{fig:robustVAE}, we added some noise to the latent vectors obtained from the training data, and decoded the resulting vectors.  More precisely, the valid latent vector $\vec z$ of an input image is shifted along the axis defined by the single-neuron regression network parameters to obtain the noisy latent vector $\vec z^* = \vec z + \alpha \theta_c$.  This warped latent vector is decoded to produce a segmentation map $\mathbf{x}^* = \mbox{dec}(\vec z^*)$.  Because of the linear constraint, the distribution is stretched along a plane perpendicular to the the axis defined by the single-neuron regression network, to allow for a linear separation of the domain-specifc target. At equal magnitude, warping the latent vector along the normal of the plane that defines the stretch of the distribution is more likely to produce samples out of distribution than along any other direction in the latent space.  Since out-of-distribution samples are more likely to be decoded into implausible segmentation maps, this perturbation of the latent vector is a suitable way to obtain an artificial anatomically invalid cardiac shape paired with the original valid cardiac shape.

The rVAE is trained to recreate $\mathbf{x}$ from $\mathbf{x}^*$. An additional constraint is used to incite the encoder to project erroneous segmentations close to their corresponding valid latent vectors. This constraint is implemented as an additional KL loss term. This KL loss term minimizes the distance between the latent vector obtained by the rVAE on the noisy data and the original latent vector generated by the cVAE on the clean training data. 

For a given $\mathbf{x}$, $\mathbf{x}^*$ and $t$  the loss function is:

\begin{eqnarray}
I\!\!E_{q_{\theta_e'}(\vec z|\mathbf{x}^*)}[-\log q_{\theta_d'}(\mathbf{x}|\vec z)]+KL(p_{\theta_e'}(\vec z|\mathbf{x}) \| p(\vec z)) + \\ 
\|y_{\theta_c'}(\vec z)-t\|^2 + KL(p_{\theta_e'}(\vec z|\mathbf{x}^*) \| p_{\theta_e}(\vec z|\mathbf{x})) \nonumber
\end{eqnarray}

\begin{table}[tp]
    \centering
	\caption{Ablation study of our cVAE showing the average \% of anatomical errors while navigating through the latent space.}
	\begin{tabular*}{0.49\textwidth}{@{\extracolsep{\fill} } l ccccc}
    	\toprule
    	\multirow{2}[3]{*}{Dataset} & \multicolumn{1}{c}{AE} & \multicolumn{4}{c}{VAE} \\
    	\cmidrule(lr){2-2} \cmidrule(lr){3-6}
    	&& \mcc{-} & \mcc{Registered} & \mcc{Const.} & \mcc{Reg. + const.} \\
    	\midrule
        ACDC  & 64.76 & 5.84 & 5.85 & 8.48 & 1.25 \\ 
    	CAMUS & 41.25 & 1.48 & 0.52 & 3.32 & 0.12 \\
    	\bottomrule
	\end{tabular*}
\label{tab:ablationStudyVAE}
\end{table}

\subsection{Implementation Details}\label{sec:implementation}
The {\em encoder} of our cVAE is made up of 4 convolutional blocks, followed by two fully-connected heads that output the $\vec \mu$ and $\Sigma$ parameters of the posterior distribution. Each convolutional block consists of two $3 \times 3$ convolutional layers with ELU~\cite{Clevert2015FastAA} activations: the first one with stride 2 (to downsample by half in lieu of pooling), and the second with stride 1 and same padding. The dimensionality of the latent space was fixed at 32, to remain as low as possible while allowing for high reconstruction accuracy.

The {\em decoder} follows a similar structure, first using a fully-connected layer to project to the same volume as the output of the last convolutional block in the encoder. After the FC layer comes a 4-block structure mirroring the encoder. Each block now consists of 2 layers with ELU~\cite{Clevert2015FastAA} activations: the first one is a $2 \times 2$ transposed convolution with stride 2 (to upsample by 2), and the second one is a $3 \times 3$ convolution with stride 1 and same padding. A final $3 \times 3$ convolution layer with stride 1 and same padding outputs the pixel-wise score for each class.

The number of feature maps is set to 48 for the first layer, and doubles at each successive block in the encoder. It follows the reverse logic in the decoder, where it is reduced by half in each block in order to reach 48 just before the final convolution with softmax. The encoder and decoder are trained end-to-end with the Adam optimizer~\cite{kingma2014adam}, using a learning rate of $6\times10^{-5}$ for ACDC and $5\times 10^{-4}$ for CAMUS. In both cases, a $L2$ weight regularization with $\lambda = 0.01$ was applied.

The AE mentioned in the ablation study of Table~\ref{tab:ablationStudyVAE} uses the exact same architecture and hyperparameter values, except for one adaptation. At the end of the encoder, a single fully-connected head is used to directly obtain the latent vector, instead of the parameters of the posterior distribution.

The segmentations maps were resized to $256\times 256$ and registered. In the case of ACDC, the registration process implied centering image on the LV, and aligning the LV and RV on an horizontal line (i.e. aligning according to the centers of the cavities). With CAMUS, registering meant centering the image on the union of the LV and MYO, and vertically aligning the principal axis of the LV. During inference, the registration is based on the results of the segmentation method rather than the groundtruth. Because this is done prior to any of our post-processing, our method is dependent on the original segmentation being at least somewhat accurate w.r.t. the position and orientation of the heart.

\begin{table}[tp]
    \centering
	\caption{Number of anatomically invalid segmentation results on the ACDC test set (1076 images) for 11 segmentation methods with and without our post-processing methods (w/o RS: without rejection sampling, w/ RS: with rejection sampling, Dicho: dichotomic search).}
	\begin{tabular*}{0.49\textwidth}
        {@{} @{\extracolsep{\fill}} p{0.08\textwidth} c cc ccc @{}}
		\toprule
		\multirow{2}[3]{*}{Methods} & \multicolumn{1}{c}{Original}
        & \multicolumn{2}{c}{VAE} & \multicolumn{3}{c}{Nearest Neighbors} \\
		\cmidrule(lr){2-2} \cmidrule(lr){3-4} \cmidrule(lr){5-7}
		&& \mcc{-} & \mcc{Robust} & \mcc{w/o RS} & \mcc{w/ RS} & \mcc{Dicho} \\
		\midrule
        Zotti-2~\cite{Zotti18}      & 55  & 16  & 7 & 0 & 0 & 0 \\
        Khened~\cite{khened2017densely}       & 55  & 16  & 9 & 0 & 0 & 0 \\
        Baumgartner~\cite{baumgartner2017exploration}  & 79  & 17  & 8 & 0 & 0 & 0 \\
		Zotti~\cite{zotti2017gridnet}        & 82  & 15  & 7 & 0 & 0 & 0 \\
        Grinias~\cite{grinias2017fast}      & 89  & 12  & 6 & 0 & 0 & 0 \\
        Isensee~\cite{isensee2017automatic}      & 128 & 21  & 7 & 0 & 0 & 0 \\
        Roh{\'e}~\cite{rohe2017automatic}         & 287 & 40  & 21 & 0 & 0 & 0 \\
        Wolterink~\cite{wolterink2017automatic}    & 324 & 42  & 16 & 0 & 0 & 0 \\
        Jain~\cite{patravali20172d}         & 185 & 28  & 17 & 0 & 0 & 0 \\
        Yang~\cite{jang2017automatic}         & 572 & 182 & 137 & 0 & 0 & 0 \\
        ACNN~\cite{Oktay17}         & 139 & 41  & 21 & 0 & 0 & 0 \\ 
		\bottomrule
	\end{tabular*}
\label{tab:AcdcAblationStudyAnatomicalMetrics}
\end{table}

\section{Experimental Setup and Results}\label{sec:setup}

\subsection{Datasets, evaluation criteria, and other methods}\label{sec:data}

\subsubsection{MRI dataset}
The MRI dataset is the 2017 ACDC dataset~\cite{Bernard2018DeepLT}, which contains short-axis cine-MR images of 150 patients: 100 for training and 50 for testing.  Particularly, a series of short axis slices cover the LV from the base to the apex, with one image every 5 or 10 mm, according to the examination. The spatial resolution goes from 1.37 to 1.68 mm\textsuperscript{2}/pixel and 28 to 40 images cover the cardiac cycle. The end-diastolic and end-systolic phases were visually selected.  As shown in Fig.~\ref{fig:result}(a), the LV, RV and MYO of every patient has been manually segmented.  We report the average  Hausdorff distance (HD) and 3D Dice index for the LV, RV and MYO as well as the LV and RV ejection fraction (EF) absolute error.  Since our approach can post-process any segmentation method, we tested it on the test results reported by ten ACDC challengers.  Their methods are summarized by Bernard \mbox{\emph{et al.}}~\cite{Bernard2018DeepLT} except for  Zotti-2~\cite{Zotti18} whose results have been uploaded after the challenge.  We also report results for the ACNN method of Oktay \mbox{\emph{et al.}}~\cite{Oktay17} that uses a latent anatomical prior together with their segmentation CNN. Results from our best ACNN implementation (which involves a U-Net and our VAE) are very close to that of the original paper, despite the fact that the ACDC training set is smaller than in the original paper~\cite{Oktay17}.  HD values are also slightly larger since we use a 3D HD instead of a 2D HD as in the original paper.

\subsubsection{Echocardiographic dataset}
The CAMUS dataset~\cite{Leclerc19} consists of conventional clinical exams from 500 patients acquired with a GE Vivid E95 ultrasound scanner. The acquisitions were optimized to perform measurements of the left ventricular ejection fraction. For each patient, 2D apical four-chamber and two-chamber view sequences were acquired with the same acquisition protocol and exported from EchoPAC analysis software (GE Vingmed Ultrasound, Horten, Norway). The corresponding videos are expressed in native polar coordinates. The same resampling scheme was applied on each sequence to express the corresponding images into a cartesian coordinate system with a constant grid resolution of $\lambda/2$ (i.e. 0.31 mm) in the lateral direction and $\lambda/4$ (i.e. 0.15 mm) in the axial direction, where $\lambda$ corresponds to the wavelength of the ultrasound probe.  The dataset is divided in 10 folds of equal size, nine of which are used for training and one for testing.  The image quality (poor, good, and medium) and ejection fraction ($\leq$ 45\%, $\geq$ 55\% or in between) are uniformly distributed across every fold.  A senior cardiologist manually annotated the endocardium and epicardium borders of the left ventricle as well as the atrium of the end-diastolic (ED) and end-systolic (ES) images of every patient.

We tested our framework on the output of 7 methods:~four conv nets (U-Net~\cite{Ronneberger2015UNet, Leclerc19}, LUNet, ENet~\cite{Paszke2016ENetAD} and SHG~\cite{Newell2016SHG}) and three non-deep learning methods (SRF~\cite{Dollar2015SRF}, BEASM-auto~\cite{Pedrosa2017BEASM, Barbosa2013BEASM-auto}, and BEASM-semi~\cite{Pedrosa2017BEASM, Leclerc19}).  Note the non-deep-learning methods were state-of-the-art up until 2017.

\begin{table*}[tp]
    \centering
    \caption{Accuracy and clinical metrics of SOTA segmentation methods, with and without our post-processing, on the ACDC test set. [Top] Average Dice index and Hausdorff distance (in mm). [Bottom] Average error (in \%) on LV and RV ejection fraction (EF).}
    \smallskip
    \begin{tabular*}{0.75\textwidth}
        {@{} @{\extracolsep{\fill}} l c cc ccc @{}}
        \toprule
        \multirow{2}[3]{*}{Methods} & \multicolumn{1}{c}{Original}
        & \multicolumn{2}{c}{VAE} & \multicolumn{3}{c}{Nearest Neighbors} \\
		\cmidrule(lr){2-2} \cmidrule(lr){3-4} \cmidrule(lr){5-7}
		&& \mcc{-} & \mcc{Robust} & \mcc{w/o RS} & \mcc{w/ RS} & \mcc{Dicho} \\
        \midrule
        Zotti-2~\cite{Zotti18}      & .913 / 9.7  & .910 / 10.1 & .910 / 11.3 & .899 / 14.4 & .909 / 11.0 & .910 / 10.1 \\
        Khened~\cite{khened2017densely}       & .915 / 11.3 & .912 / 12.3 & .912 / 11.8 & .894 / 15.2 & .909 / 12.7 & .912 / 10.9 \\
        Baumgartner~\cite{baumgartner2017exploration}  & .914 / 10.5 & .911 / 11.2 & .912 / 10.8 & .889 / 18.2 & .907 / 12.6 & .910 / 10.6 \\
        Zotti~\cite{zotti2017gridnet}        & .910 / 9.7  & .907 / 10.9 & .907 / 11.3 & .878 / 19.6 & .903 / 12.6 & .907 / 11.0 \\
        Grinias~\cite{grinias2017fast}      & .835 / 15.9 & .833 / 19.3 & .834 / 15.7 & .752 / 32.5 & .825 / 16.9 & .833 / 15.8 \\
        Isensee~\cite{isensee2017automatic}      & .926 / 9.1  & .923 / 10.7 & .923 / 9.7 & .881 / 18.4 & .917 / 11.2 & .923 / 9.2  \\
        Roh{\'e}~\cite{rohe2017automatic}         & .891 / 12.2 & .887 / 14.6 & .886 / 16.3 & .756 / 32.2 & .874 / 15.1 & .887 / 12.8 \\
        Wolterink~\cite{wolterink2017automatic}    & .907 / 10.8 & .903 / 13.0 & .902 / 11.6 & .752 / 32.8 & .887 / 13.5 & .903 / 11.0 \\
        Jain~\cite{patravali20172d}         & .891 / 12.2 & .886 / 12.6 & .885 / 13.0 & .820 / 31.9 & .878 / 14.2 & .886 / 11.6 \\
        Yang~\cite{jang2017automatic}         & .800 / 27.5 & .752 / 21.7 & .742 / 24.3 & .455 / 29.7 & .722 / 11.5 & .752 / 10.2 \\
        ACNN~\cite{Oktay17}         & .892 / 12.3 & .886 / 26.2 & .885 / 21.6 & .885 / 12.0 & .885 / 12.2 & .889 / 13.1\\ 
        \bottomrule 
        \toprule
        Zotti-2~\cite{Zotti18}      & 2.54  /  5.11 & 2.63 / 5.12 & 2.59 / 5.08 & 2.49 / 5.57 & 2.58 / 5.18 & 2.62 / 5.18 \\
        Khened~\cite{khened2017densely}       & 2.39 / 5.24 & 2.41 / 4.96 & 2.43 / 5.18 & 2.70 / 5.36 & 2.63 / 5.07 & 2.42 / 5.27 \\
        Baumgartner~\cite{baumgartner2017exploration}  & 2.58 / 6.00 & 2.62 / 6.30 & 2.54 / 6.18 & 2.83 / 6.72 & 2.85 / 6.48 & 2.64 / 6.33 \\
        Zotti~\cite{zotti2017gridnet}        & 2.98 / 5.48 & 2.98 / 5.42 & 3.04 / 5.57 & 3.06 / 5.72 & 3.10 / 5.71 & 3.06 / 5.59 \\
        Grinias~\cite{grinias2017fast}      & 4.14 / 7.39 & 4.18 / 7.86 & 3.94 / 7.59 & 4.67 / 8.00 & 4.33 / 7.35 & 4.01 / 7.43 \\
        Isensee~\cite{isensee2017automatic}      & 2.16 / 4.85 & 2.15 / 4.61 & 2.18 / 4.85 & 2.49 / 5.58 & 2.35 / 4.48 & 2.20 / 4.82 \\
        Roh{\'e}~\cite{rohe2017automatic}         & 2.84 / 8.18 & 2.95 / 7.85 & 2.85 / 8.34 & 3.13 / 8.93 & 3.39 / 7.97 & 2.91 / 8.11 \\
        Wolterink~\cite{wolterink2017automatic}    & 2.75 / 6.59 & 2.82 / 6.39 & 2.83 / 6.42 & 3.40 / 6.93 & 3.48 / 6.07 & 2.84 / 6.44 \\
        Jain~\cite{patravali20172d}         & 4.36 / 8.49 & 4.35 / 8.83 & 4.46 / 9.09 & 4.98 / 9.63 & 4.59 / 8.69 & 4.40 / 8.72 \\
        Yang~\cite{jang2017automatic}         & 6.22 / 15.99 & 6.80 / 20.56 & 5.40 / 21.58 & 7.57 / 27.9 & 7.77 / 22.09 & 9.10 / 21.76 \\
        ACNN~\cite{Oktay17}         & 2.46 / 3.68 & 2.53 / 4.09 & 2.59 / 4.05 & 2.51 / 3.89 & 2.96 / 3.82 & 2.50 / 3.71 \\ 
        \bottomrule
    \end{tabular*}
\label{tab:AcdcAblationStudySimilarityMetrics}
\end{table*}

\subsection{Experimental Results}\label{sec:results}

\subsubsection{Constrained variational autoencoder}
\begin{figure}[tp]
\centering
  \begin{subfigure}[b]{\columnwidth}
    \includegraphics[width=\textwidth]{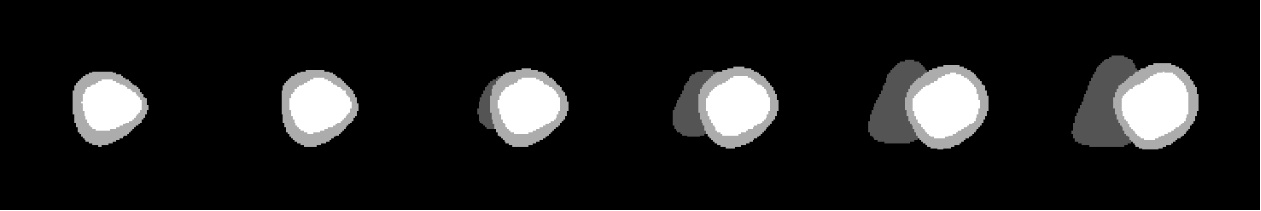}
    \caption{MRI (ACDC)}
  \end{subfigure}
  \vfill
  \begin{subfigure}[b]{\columnwidth}
    \includegraphics[width=\textwidth]{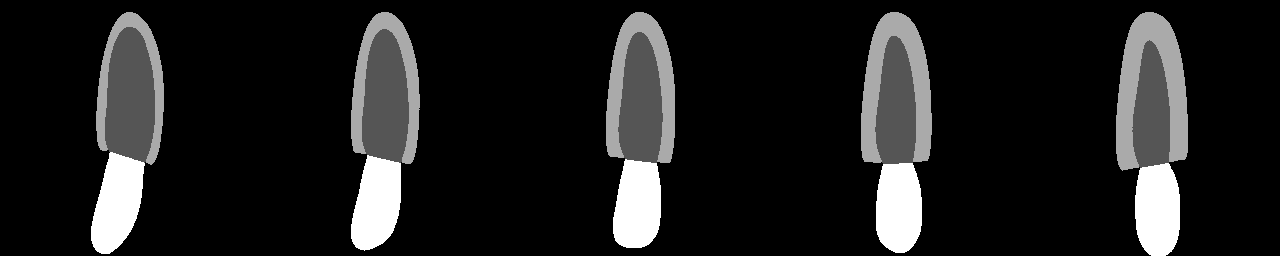}
    \caption{US (CAMUS)}
  \end{subfigure}
  \caption{\small  The left-most and right-most images are groundtruth configurations for both MRI and US while the other configurations were obtained though a linear interpolation. \vspace{-0.3cm}}
  \label{fig:interpolation}
\end{figure}

We gauged the linearity property of the latent space generated by our cVAE through the ablation study in  Table~\ref{tab:ablationStudyVAE}.  Since our post-processing method relies on latent vector interpolation (c.f. Eq~(\ref{eq:optim_dicho})), we computed the percentage of anatomically incorrect results obtained after interpolating a series of two valid latent vectors chosen at random.  To do so, we iteratively selected two  random groundtruth images from two random patients, projected it to the latent space with the cVAE encoder and linearly interpolated 25 new latent vectors. We then converted these 25 vectors back into the image space with the cVAE decoder and computed their percentage of anatomical errors.  This procedure is illustrated in Fig.~\ref{fig:interpolation}.

We repeated that process 300 times, i.e. combinations between 25 random vectors, (both for Camus and ACDC) for the cVAE with and without registration and with and without the one-neuron regression net.   We first tested our full method (i.e. with image registration and a L2 regression constraint), then removed the image registration but kept the regression constraint, then removed the regression and kept the image registration and finally only used the VAE without registration nor regression.
As shown in Table~\ref{tab:ablationStudyVAE}, the combination of image registration and regression constraint reduces the percentage of anatomically implausible results down to $1.25\%$ for ACDC and a negligeable $0.12\%$ for CAMUS, which is more than 4x lower than for any other configuration.  As for a simple autoencoder (c.f the AE column), since it provides no constraint on the latent space whatsoever, the percentage of errors is orders of magnitude larger.
\vspace{-0.0cm}

\subsubsection{ACDC Postprocessing results}
Results on the ACDC test set are in Table~\ref{tab:AcdcAblationStudyAnatomicalMetrics} and Table~\ref{tab:AcdcAblationStudySimilarityMetrics}. Table~\ref{tab:AcdcAblationStudyAnatomicalMetrics} contains the total number of slices with at least one anatomical error for 11 different methods, the first ten being official ACDC challengers. Results without our post-processing are under the {\em Original} column.  As for Table~\ref{tab:AcdcAblationStudySimilarityMetrics}, we report for the same methods their associated Dice index and Hausdorff distance (HD) [top] as well as their LV and RV ejection fraction absolute error [bottom].

As can be seen from the "VAE" column in Table~\ref{tab:AcdcAblationStudyAnatomicalMetrics}, feeding every erroneous segmentation map to our VAE without transforming the latent vector $\vec z$, significantly reduces the number of anatomical errors, without affecting too much the average clinical metrics (Table~\ref{tab:AcdcAblationStudySimilarityMetrics}). This comes as no surprise, since the VAE was trained to output similar anatomically correct cardiac shapes (the ACDC test set has a total of $1078$ slices).  The rVAE further reduces, by a factor of almost 2, the number of anatomical errors, without significantly impacting the overall anatomical metrics.  With a processing time 10 times faster than our most accurate method, the rVAE can be seen as a good compromise for real-time applications.

However, like any neural network, a VAE (be it robust or not) comes with no guarantee on the validity of its output.  To completely eliminate erroneous segmentations, we tested three variants of our method.  At first, we swap erroneous latent vectors with their nearest neighbor (thus forcing $\alpha=1$ in Eq.~\ref{eq:optim_dicho}) without and with rejection sampling (cf. columns "w/o RS" and "w/ RS").  As mentioned in Section~\ref{sec:augmentation}, we increased to 4 million the number of anatomically correct latent vectors with the rejection sampling.  Despite the fact that both methods reduce to zero the number of anatomical errors, we can see from Table~\ref{tab:AcdcAblationStudySimilarityMetrics} that data augmentation systematically produces better results.  Also, while the improvements are incremental for top performing methods (e.g. the Dice index of Zotti-2 went from .899 to .909 and its HD from 14.4 to 11.0), they are drastic for methods with a large number of anatomical errors (e.g. Wolterink saw its Dice index go from .752 to .887 and its HD from 32.2 to 13.5).  We can thus conclude that our method without a data augmented latent space could hurt the overall accuracy of certain methods.

\begin{table}[tp]
    \centering
	\caption{Number of anatomically invalid segmentation results in cross-validation on CAMUS (2000 images) for 7 segmentation methods with and without our post-processing methods. The methods in the upper half segment all 3 classes (LV\textsubscript{endo}, LV\textsubscript{epi} and LA), whereas the methods in the lower half only segment 2 classes (LV\textsubscript{endo} and LV\textsubscript{epi}).}
	\smallskip
	\resizebox{0.48\textwidth}{!}{
	\begin{tabular*}{0.49\textwidth}
        {@{} @{\extracolsep{\fill}} p{0.135\textwidth} c cc ccc @{}}
		\toprule
		\multirow{2}[3]{*}{Methods} & \multicolumn{1}{c}{Original}
        & \multicolumn{2}{c}{VAE} & \multicolumn{3}{c}{Nearest Neighbors} \\
		\cmidrule(lr){2-2} \cmidrule(lr){3-4} \cmidrule(lr){5-7}
		&& \mcc{-} & \mcc{Robust} & \mcc{w/o RS} & \mcc{w/ RS} & \mcc{Dicho} \\
		\midrule
        U-Net~\cite{Ronneberger2015UNet, Leclerc19}       & 84 & 16 & 14 & 0 & 0 & 0 \\
        LUNet~\cite{Leclerc2020Lunet}       & 25 & 11 & 6  & 0 & 0 & 0 \\
        ENet~\cite{Paszke2016ENetAD}        & 69 & 21 & 22 & 0 & 0 & 0 \\
        SHG~\cite{Newell2016SHG}         & 38 & 5  & 5  & 0 & 0 & 0 \\ \hline
        SRF~\cite{Dollar2015SRF}         & 101 & 46 & 48 & 1 & 2 & 2 \\
        {\scriptsize BEASM-auto ~\cite{Pedrosa2017BEASM, Barbosa2013BEASM-auto}}  & 12 & 2  & 3  & 0 & 0 & 0 \\
		{\scriptsize BEASM-semi~\cite{Pedrosa2017BEASM, Leclerc19}}  & 10 & 4  & 7  & 0 & 0 & 0 \\ 
		\bottomrule
	\end{tabular*}
	}
\label{tab:CamusAblationStudyAnatomicalMetrics}
\end{table}

The last column of Tables~\ref{tab:AcdcAblationStudyAnatomicalMetrics} and ~\ref{tab:AcdcAblationStudySimilarityMetrics} shows the results of our complete method, i.e.  Eq.(~\ref{eq:optim_dicho}) optimized with a dichotomic search on a data augmented latent space.   While all results respect the anatomical criteria, the EF error and the Dice index are almost identical to that of the original methods.  The HD also never increases more than 1.3 mm. Considering that the average voxel size is near 1.4x1.4x10 mm\textsuperscript{3}, the increase corresponds to less than 1 pixel in the image.  This shows that our approach does not degrade the overall results of a given approach.

\begin{figure*}[tp]
\centering
  \begin{subfigure}[b]{0.48\textwidth}
    \includegraphics[width=\textwidth]{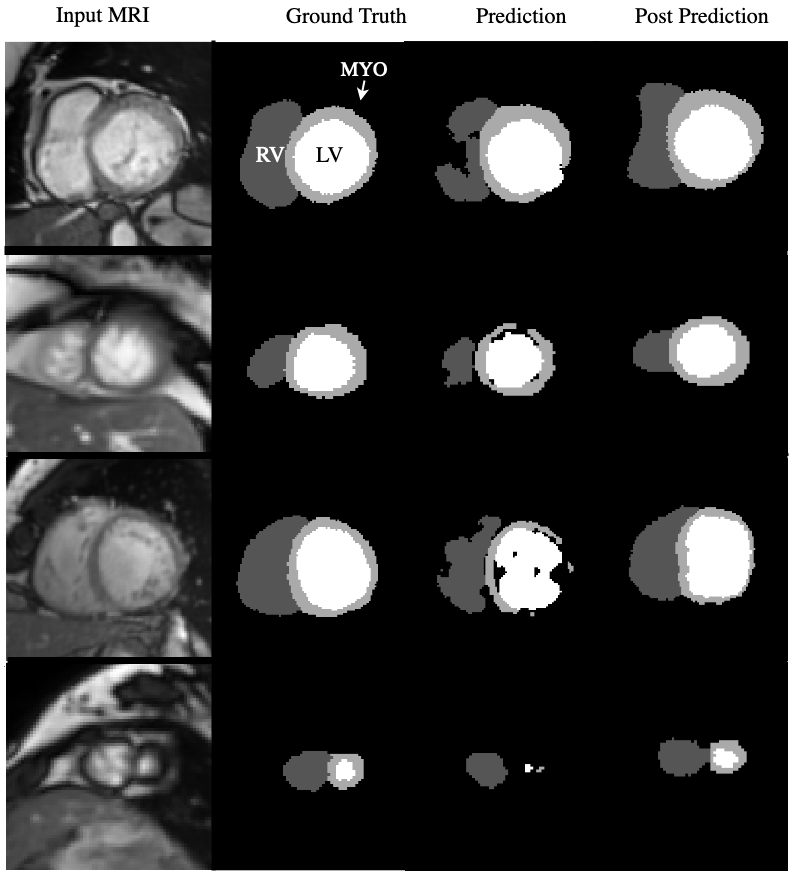}
    \caption{MRI (ACDC)}
  \end{subfigure}
  \begin{subfigure}[b]{0.5\textwidth}
    \includegraphics[width=\textwidth]{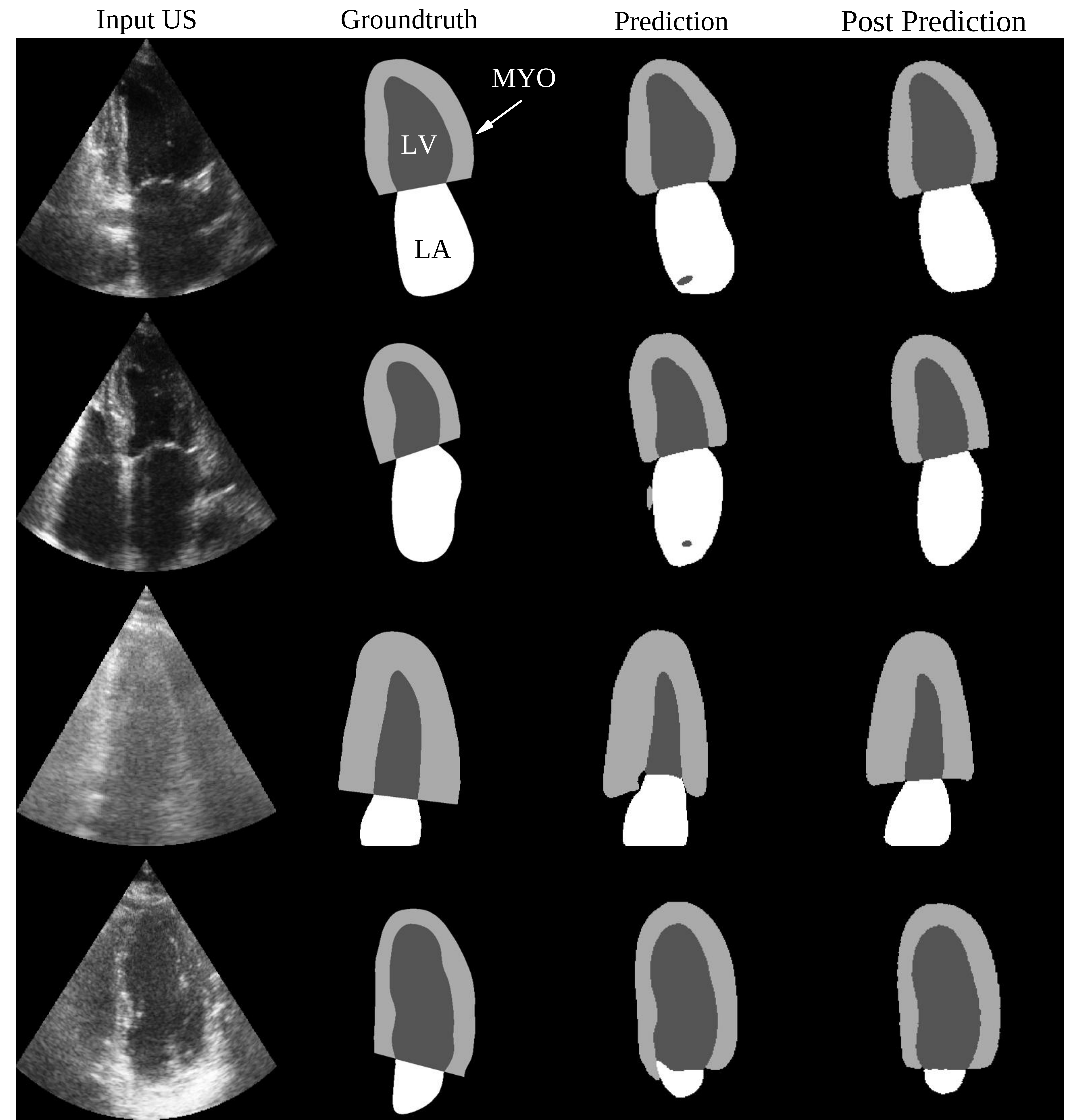}
    \caption{US (CAMUS)}
  \end{subfigure}
  \caption{\small Groundtruth and erroneous maps before and after our post-processing method.\vspace{-0.5cm}}
  \label{fig:result}
\end{figure*}

Fig.~\ref{fig:result}~(a) shows erroneous predictions before and after our post-processing.  While the correct areas are barely affected by our method, erroneous sections, big or small, get smoothly warped.  Our method takes roughly 1 sec to process a 2D image on a mid-end computer equipped with a Titan X GPU.

\begin{table*}[tp]
    \centering
	\caption{Accuracy and clinical metrics of SOTA segmentation methods, with and without our post-processing, in cross-validation on CAMUS dataset. [Top] Average Dice index and Hausdorff distance (in mm). [Bottom] Average error (in \%) on LV ejection fraction (EF).}
	\smallskip
	\begin{tabular*}{0.8\textwidth}
        {@{} @{\extracolsep{\fill}} l cc cc ccc @{}}
		\toprule
		\multirow{2}[3]{*}{Methods} & \multicolumn{1}{c}{Original}
        & \multicolumn{2}{c}{VAE} & \multicolumn{3}{c}{Nearest Neighbors} \\
		\cmidrule(lr){2-2} \cmidrule(lr){3-4} \cmidrule(lr){5-7}
		&& \mcc{-} & \mcc{Robust} & \mcc{w/o RS} & \mcc{w/ RS} & \mcc{Dicho} \\
		\midrule
        U-Net~\cite{Ronneberger2015UNet, Leclerc19}       & .921 / 6.0  & .923 / 5.7  & .923 / 5.7  & .922 / 5.7  & .922 / 5.7  &  .923 / 5.7  \\
        LUNet~\cite{Leclerc2020Lunet}       & .922 / 5.9  & .921 / 5.9  & .922 / 5.9  & .921 / 5.9  & .921 / 6.0  & .921 / 6.0  \\
        ENet~\cite{Paszke2016ENetAD}        & .923 / 5.8  & .921 / 5.9  & .921 / 5.9  & .920 / 5.9  & .920 / 5.9  & .921 / 5.9  \\
        SHG~\cite{Newell2016SHG}         & .915 / 6.2  & .915 / 6.2  & .916 / 6.2  & .915 / 6.2  & .915 / 6.2  & .915 / 6.2  \\ \hline
        SRF~\cite{Dollar2015SRF}         & .879 / 13.1 & .877 / 13.2 & .878 / 13.2 & .879 / 13.0 & .879 / 13.0 & .879 / 13.0  \\
        BEASM-auto~\cite{Pedrosa2017BEASM, Barbosa2013BEASM-auto}  & .868 / 10.5 & .868 / 10.5 & .867 / 10.5 & .868 / 10.5 & .868 / 10.5 & .868 / 10.5 \\
		BEASM-semi~\cite{Pedrosa2017BEASM, Leclerc19}  & .899 / 7.8  & .899 / 7.8  & .899 / 7.8  & .899 / 7.8  & .899 / 7.8  & .899 / 7.8  \\ 
		\bottomrule 
        \toprule
        U-Net~\cite{Ronneberger2015UNet, Leclerc19}       & 5.4  & 5.6  & 5.6  & 5.9  & 5.9  & 5.7  \\
        LUNet~\cite{Leclerc2020Lunet}       & 5.1  & 5.1  & 5.1  & 5.4  & 5.2  & 5.2  \\
        ENet~\cite{Paszke2016ENetAD}        & 5.6  & 5.4  & 5.4  & 5.5  & 5.6  & 5.4  \\
        SHG~\cite{Newell2016SHG}         & 5.8  & 5.9  & 5.9  & 6.1  & 6.1  & 6.0  \\ \hline
        SRF~\cite{Dollar2015SRF}         & 12.7 & 14.5 & 14.3 & 14.4 & 14.4 & 14.3 \\
        BEASM-auto~\cite{Pedrosa2017BEASM, Barbosa2013BEASM-auto}  & 10.5 & 10.5 & 10.5 & 10.6 & 10.5 & 10.5 \\
		BEASM-semi~\cite{Pedrosa2017BEASM, Leclerc19}  & 9.8  & 9.8  & 9.8  & 9.8  & 9.8  & 9.8  \\ 
	\end{tabular*}
\label{tab:CamusAblationStudySimilaritylMetrics}
\end{table*}

\subsubsection{CAMUS Postprocessing results}
We  perform a similar set of experiments on the CAMUS dataset.  Results are reported in Table~\ref{tab:CamusAblationStudyAnatomicalMetrics} (number of anatomically invalid slices) and Table ~\ref{tab:CamusAblationStudySimilaritylMetrics} (clinical metrics).  As for ACDC, the use of a simple VAE significantly reduces the number of anatomical errors, without affecting too much the average Dice index, HD and EF absolute error.  However, unlike for ACDC, the robust VAE did not succeed at further reducing errors, especially for the non-deep-learning methods.  This may be explained by the fact that the number of anatomical errors are already low with a basic VAE. 

Another difference with ACDC is the results for our three nearest neighbors methods.  While they reduce to zero the number of anatomical errors, all three methods have almost the same Dice index, HD and anatomical errors.  This can be explained by the fact that the long-axis cardiac shapes are roughly similar from one patient to another, regardless of the time instant (c.f.~Fig.~\ref{fig:interpolation}).  This is unlike the short-axis view, where the shape varies greatly between the basis of the heart down to the apex. As such, the long-axis valid latent vectors are probably closer together, so a simple nearest neighbor swap is enough to enforce our anatomical criteria on the output while preserving the overall anatomical shape.  

However, as can be seen from Table~\ref{tab:CamusAblationStudySimilaritylMetrics}, like for ACDC, our method does not degrade by a significant manner the anatomical nor the clinical metrics.

\subsubsection{Inter-observer variability}
The inter-observer variability of cardiac MRI and echocardiographic image segmentation was reported by Bernard \mbox{\emph{et al.}}\cite{Bernard2018DeepLT} and Leclerc \mbox{\emph{et al.}}\cite{Leclerc19}.  For MRI segmentation, on average the inter-observer Dice score for the LV, the RV and the MYO at the end-systolic and end-diastolic time instant is $0.90$ while the average Hausdorff distance is $9.3$ mm~\cite{Bernard2018DeepLT} .  As can be seen from Table~\ref{tab:AcdcAblationStudySimilarityMetrics}, the methods with a dice score above $0.90$ (column $Original$) are also above $0.90$ after our processing (column $Dicho$).  Also, the only method with a Hausdorff distance below $9.3$ mm is that of Isensee~\cite{isensee2017automatic}, which is also below $9.3$ after our processing.

As for the echocardiographic segmentation, the average inter-observer Dice score reported in \cite{Leclerc19} is $0.899$ and the average Hausdorff distance is $7.34$ mm.  Again, as can be seen from the $Original$ column of Table~\ref{tab:CamusAblationStudySimilaritylMetrics}, the first four methods are within the observer-variability, and still are after our post-processing (column $Dicho$).

This reveals that while our method guarantees to produce results that follow pre-defined anatomical guidelines, it does not degrade the overall accuracy of highly effective methods.

\subsubsection{Post-processsing degenerated results}
\begin{figure}[tp]
\centering
  \begin{subfigure}[b]{\columnwidth}
    \includegraphics[width=\textwidth]{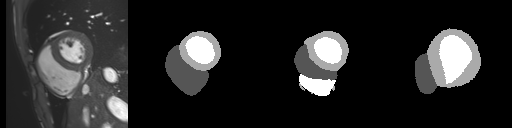}
    \caption{Valid but inaccurate post-processed segmentation for MRI}
  \end{subfigure}
  \vfill
  \begin{subfigure}[b]{\columnwidth}
    \includegraphics[width=\textwidth]{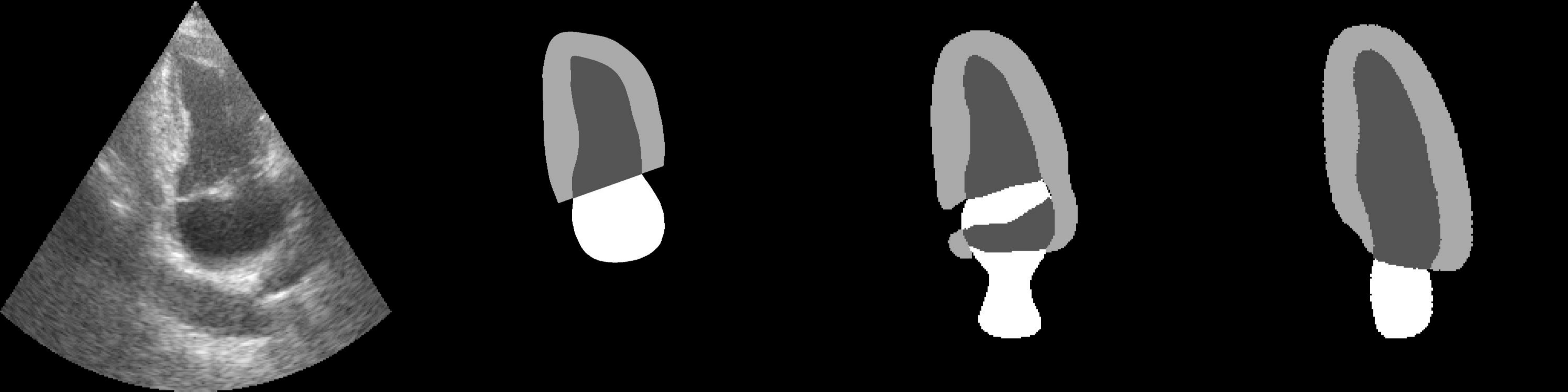}
    \caption{Valid but inaccurate post-processed segmentation for US}
  \end{subfigure}
  \vfill
  \begin{subfigure}[b]{\columnwidth}
    \includegraphics[width=\textwidth]{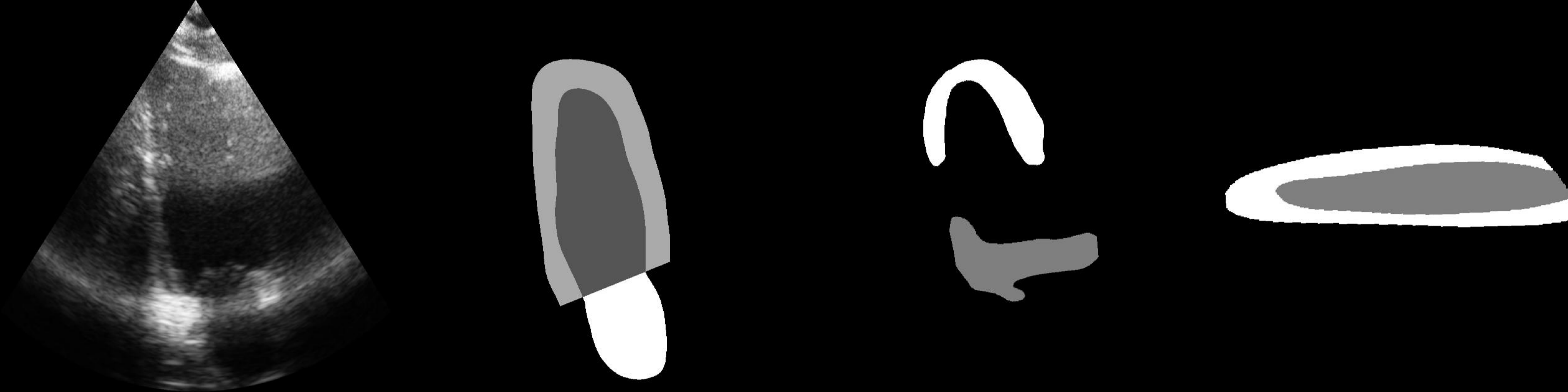}
    \caption{Invalid post-processed segmentation for US}
  \end{subfigure}
  \caption{\small Groundtruth and erroneous maps before and after our post-processing method, when original segmentations are degenerated. From left to right, the images in all 3 figures are: input (MRI or US), groundtruth, original prediction, prediction after anatomical postprocessing. \vspace{-0.6cm}}
  \label{fig:degenerated_result}
\end{figure}

Our method has its own limits and cannot be regarded as a solution to every harm.  While our method guarantees the anatomical validity, w.r.t. the hardcoded criteria, of the output, it by no means guarantees that the produced output is close to the groundtruth.  As such, if the erroneous segmentation map $\mathbf{x}'$ it has to correct has little to no overlap with the groundtruth, our method will not necessarily warp $\mathbf{x}'$ in the direction of the groundtruth.  It will only warp $\mathbf{x}'$ to the closest correct cardiac shape.  Three such examples are provided in Fig.~\ref{fig:degenerated_result} where the result of our method is not closer to the groundtruth than $\mathbf{x}'$.  In fact, the cardiac shape of Fig.~\ref{fig:degenerated_result}(c) is so degenerated that the produced output is perpendicular to the groundtruth (because the inverse registering operation is based on the principal axis of the LV, which in this case is horizontal).  Also, despite the fact that the produced shape is anatomically valid, the segmentation is sideways, causing the computation for the LV's width and MYO's thickness (c.f. criterion 6 in sec.~\ref{sec:criteria:la}) to be inaccurate and to detect an anomaly, hence the 1 and 2 errors reported for the SRF method in Table~\ref{tab:CamusAblationStudyAnatomicalMetrics}.  This particular example also illustrates what can happen when the original segmentation is so bad that even the inference-time registering is inaccurate, as mentioned at the end of sec.~\ref{sec:implementation}.  That said, even for inaccurate segmentation methods (e.g. Grinias and Yang in Table~\ref{tab:AcdcAblationStudySimilarityMetrics}), our method does not worsen their overall scores.  Metrics obtained solely on the anatomically incorrect images are provided in the supplementary materials and also show that our method does not reduce the overall metrics.

\vspace{-0.0cm}
\section{Conclusion}\label{sec:conclusion}
\vspace{-0.2cm}
We proposed a post-processing cVAE that converts invalid cardiac shapes into close but correct shapes.  This is done by replacing the latent vector of an invalid shape by a close but valid latent vector.   Intensive tests performed on the output of 18 segmentation methods reveal that our method is effective on both short-axis views from MRI as well as on long-axis views from US.  Our method relies on a series of anatomical criteria (16 for SA and 12 for LA) that we use both to detect abnormalities and populate a cVAE latent space.  One appealing feature of the proposed framework is that anatomical criteria do not need to be differentiable as they are not included in the loss. Furthermore, it has been shown that the warping of the incorrect segmentation shapes did not change significantly the overall geometrical metrics (Dice index and Hausdorff) nor the clinical metrics (the RV and LV ejection fraction).  As such, according to the inter and intra-expert variations reported by Bernard \mbox{\emph{et al.}}\cite{Bernard2018DeepLT} and Leclerc \mbox{\emph{et al.}}\cite{Leclerc19}, methods such as Isensee, Zotti-2, Khened and Baumgartner for ACDC and LUNet for CAMUS are within the inter-expert variation and, with our method, are now guaranteed to produce results that follow anatomical guidelines defined by the user. From the point of view of a clinical expert, it is preferable to have a plausible segmentation close to the expected one than an efficient system that spuriously provide aberrant segmentations. In that case, users cannot trust the provided physiological parameters that is calculated from these latest data, even if implausible segmentations do not significantly change the parameter values.

\vspace{-0.4cm}

\bibliographystyle{IEEEtran}
\bibliography{TMI-2020-0585}

\end{document}